\documentclass[letterpaper]{article}
\usepackage[]{aaai2027} 
\usepackage{times}
\usepackage{helvet}
\usepackage{courier}
\usepackage{xcolor}
\usepackage[hyphens]{url}
\usepackage{graphicx}
\usepackage{natbib}
\usepackage{caption}
\DeclareCaptionStyle{ruled}{labelfont=normalfont,labelsep=colon,strut=off}
\usepackage{booktabs}
\usepackage{amsmath}
\usepackage{amssymb}
\usepackage{tikz}
\usetikzlibrary{arrows.meta,positioning}
\title{NOMADD: Numerical Optimization of Models Adapting to Data Drift}
\author{Swapn Shah\textsuperscript{\rm 1}, Keith Burghardt\textsuperscript{\rm 1}}
\affiliations{
\textsuperscript{\rm 1}University of North Carolina at Charlotte\\
sshah100@charlotte.edu, keithburg@charlotte.edu
}

\begin{document}
\maketitle

\begin{abstract}
Tabular model performance degrades when feature distributions change over time or the relationship between features and outcome variables change over time, known as data drift and concept drift, respectively. These issues are challenging to mitigate in real time because labeled data may not be immediately available, or re-training a model could be impractical. While tools exist to reduce drift, they are typically bespoke to neural network architectures and adapt how models are trained. In this paper, we offer an alternative post-hoc method to reduce concept drift, which is applicable to a variety of models, from trees to neural networks to tabular foundation models. This new tool is especially useful when constraints, such as high model accuracy, bounded inference time, or model size requires users to choose between different models for their specific use-cases. Our algorithm fits the base model separately on each labeled training period, measures how its parameters evolve against a single anchor model pooled over all of those periods, compresses those changes with a low-rank factorization, and extrapolates each latent factor forward with a damped, regularized forecast. On the 18-dataset Drift-Resilient TabPFN benchmark, evaluated under that benchmark's own protocol and metric, the extrapolation improves every base family it is applied to, and achieves performance competitive with the state-of-the-art Drift-Resilient TabPFN with seconds of training. In contrast, Drift-Resilient TabPFN requires pre-training on millions of synthetic datasets over approximately 1,300 GPU-hours, and is orders of magnitude slower in inference (depending on the model). In the discussion, we explore the promise and challenges of extending this tool to other modalities.
\end{abstract}

\section{Introduction}

Tabular models are core to a range of applications, from fraud risk \cite{bao2020detecting}, to credit scoring \cite{west2000neural}, to product pricing \cite{chen2015peeking}. These models often experience issues when trained on streaming tabular data due to data drift \cite{mallick2022matchmaker,helli2024drift}, in which data features change over time, as well as the similar idea of concept drift, in which relationships between features and the outcome variable evolve over time \cite{lu2018learning}. All of these properties significantly degrade model performance, which has motivated a range of methods to detect data drift \cite{yurdakul2018statistical}, or to adapt to it, such as periodically re-training a model or online learning \cite{orabona2019modern}.

A fundamental issue, however, is that these models can be stale long before they can be re-trained. Fraud transactions \cite{Pozzolo2015}, for example, may not be known for weeks \cite{csaba2024label}, so by the time the labels exist, the newest data has already drifted. Other examples where this is common are edge devices, which have limited connectivity and therefore can only share their labels periodically. Some methods help mitigate this, such as Drift-Resilient TabPFN (DR-TabPFN) \cite{helli2024drift}, but they require heavy pre-training and are comparitively slow. They also do not extend to the smaller models most commonly deployed under regulatory, speed, or memory constraints. Banks require simple interpretable models for regulatory reasons \cite{demajo2020explainable}, and contactless payments must complete in well under a second on edge devices \cite{al2022analysing}.

To address this critical gap, we introduce NOMADD: Numerical Optimization of Models Adapting to Data Drift. Given a previously trained model, we create a separate pipeline to learn post-hoc weight changes we apply to that model. We train probing models in small time ranges that we call periods, and measure changes in the model parameters (if white box models) or log-probability fields (if foundation models) across periods. We then forecast changes in these values to anticipate what the previously trained model \emph{should} be had it been trained on later (unlabeled) data. For example, in tree ensembles, we change their leaf thresholds, while for neural networks we change the parameters themselves. 
Much like LoRA used in large language models \cite{hu2022lora}, we compress these changes with a low-rank singular value decomposition, but unlike LoRA, we then extrapolate each latent factor with a damped, ridge-regularized linear trend and apply the forecast to the anchor, shrunk by a validated factor that is allowed to be zero. 
Applying this to the benchmark datasets developed for DR-TabPFN \cite{helli2024drift}, we find that extrapolation improves base families over their frozen version, and that the improvement is largest and most reliable on lightweight models. The clearest case is XGBoost \cite{chen2016xgboost}, whose performance improves on thirteen of eighteen datasets. 
All model updates require a modicum of additional memory and achieve performance near that of foundation models with orders of magnitude less inference time.  

We summarize our contributions as follows.
\begin{itemize}
\item We create a single post-hoc procedure to reduce data and concept drift that applies to neural networks, logistic regression, tree ensembles, and foundation models.
\item We evaluate on a new benchmark and show that our method improves performance on average for each model studied.
\item When we compare our method (applied to small models) against the state-of-the-art DR-TabPFN, we find comparable performance at a fraction of its inference cost and memory (Section~\ref{sec:cost}).
\end{itemize}

\section{Related Work}

Data drift research is often split between detection and adaptation \cite{lu2018learning,gama2014survey,greco2021drift}. 
We will focus on drift adaptation methods, especially gradual data or concept drift \cite{He2021}.%

\noindent
\textbf{Model adaptation.} 
For each type of drift the easiest remedy is to re-train the model, but it is not trivial to know when is the best time to re-train, and how much history should be fed into the model. Naive re-training schedules include retraining at fixed intervals or re-training once validation performance falls below a threshold; more sophisticated methods include paired learners \cite{bach2008paired} and algorithms that learn the best training window. Recurring drift such as seasonality is better addressed with ensembles \cite{Sun2018}, and continuously evolving data is better addressed with online learning, such as the Very Fast Decision Tree Classifier \cite{domingos2000mining} or TabPFN \cite{lourencco2026context}. Importantly, none of these methods are easily adaptable to the situation where labels are delayed, which motivates NOMADD.

\noindent
\textbf{Continuous Domain Adaptation.} Adjusting models to new data is a form of continuous domain adaptation \cite{wang2020continuously,bai2023temporal} or incremental domain adaptation \cite{bitarafan2016incremental}. Continuous domain adaptation can be addressed with adversarial methods \cite{wang2020continuously,bitarafan2016incremental}, adjusting domain boundaries \cite{kumagai2016learning}, or optimal transport \cite{ortiz2020forward}. This field assumes we know both the original and the new distribution, which is infeasible when anticipating unknown future data, as is true for NOMADD.

\noindent
\textbf{Temporal domain generalization.} Temporal domain generalization methods, in contrast, predict how (neural network) models change when models experience data and concept drift. In particular, Gradient Interpolation \citep{nasery2021training} trains a neural network with a time-sensitive loss, DRAIN \citep{bai2023temporal} generates future network weights with a recurrent hypernetwork, Koodos \citep{cai2024continuous} models predict continuous parameter dynamics, and Temporal Experts Averaging \citep{tea2025} averages per-period experts with forecast weights. All are training-time frameworks tied to differentiable models. DR-TabPFN \citep{helli2024drift} meanwhile, offers an extension to the tabular foundation model, TabPFN \citep{hollmann2023tabpfn} with pretraining on second-order structural causal models whose parameters drift, unifying data and concept drift in-context. It is the published state-of-the-art on the benchmark we use.
In contrast to these previous methods, NOMADD is post hoc rather than changing how the models are trained. Moreover, it is applicable across a range of models, including tree-based models, neural networks, and foundation models. 

\section{Method}\label{sec:method}
\begin{table}[tbh!]
    \centering
    \small
    \begin{tabular}{@{}r@{\hspace{5pt}}p{0.76\columnwidth}@{}}
    \toprule
    Symbol & Description \\
    \midrule
    $M$ & Number of labeled training periods \\
    $X_m, y_m$ & Features and labels in period $m$ \\
    $g_0$ & Anchor, fit on all training periods pooled \\
    $g_m$ & Model fit on period $m$ alone \\
    $\theta(g)$ & Vector representing the boundary of $g$ \\
    $d_m$ & Boundary delta, $\theta(g_m) - \theta(g_0)$ \\
    $D$ & The $M \times P$ matrix of deltas $d_m$ \\
    $P$ & Boundary dimension, the length of $\theta$ \\
    $r$ & Rank retained by the truncated SVD of $D$ \\
    $z_{\cdot k}$ & $k$-th latent temporal trajectory \\
    $\ell_k, s_k$ & Level and slope of the fit to $z_{\cdot k}$ \\
    $h$ & Forecast horizon beyond the last period \\
    $\phi$ & Damping applied to the slope \\
    $\lambda$ & Ridge shrinkage of the slope \\
    $\alpha$ & Shrinkage applied to the forecast delta \\
    \bottomrule
    \end{tabular}
    \caption{Parameters and variables. The last four, $r$, $\phi$, $\lambda$ and $\alpha$, are chosen by forward validation inside the training periods.}
    \label{tab:params}
\end{table}
We explain how NOMADD is formulated below, with parameters defined in Table~\ref{tab:params}.
\subsection{Setting}
We observe labeled data $\{(X_m, y_m)\}_{m=0}^{M-1}$ from $M$ ordered time periods and must predict on unlabeled future periods $m \ge M$. Time periods could be pre-defined, such as days or weeks, or we can create arbitrary data bins such that a model can be trained within each bin. No future labels are available at any point, including for model selection. Let $g$ be a base learner and $\theta(g)$ be a vector that represents its parameters (or log-probability fields, if a tabular foundation model). NOMADD is post hoc in the sense that it never modifies how $g$ is trained; it only observes a sequence of ordinary fits and produces a new decision boundary for the future.

\subsection{Data restoration}
To address data drift, we are inspired by prior work on data shift~\cite{tzeng2017adversarial} to reconstruct how data would have looked were it created in the training period. Many methods could be used; indeed we explored diffusion-based methods as well as stochastic differential equation-inspired methods. We found, however, that the best method was one of the simplest. We measure the mean and the full covariance of each training data period (using Ledoit-Wolf shrinkage for stability on the small high-dimensional periods), and plot how these values change over time. We then plot these changes with linear regression, and regress each test or validation period back to the mean and covariance of the full training period. This is an optional component of our pipeline that validation selects only where it helps. It was found to never help the foundation models within the validation periods.

\subsection{Anchor and per-period deltas}\label{sec:anchor}
We fit one \emph{anchor} model $g_0$ on all training periods pooled, and one model $g_m$ on each training period $m$ separately. The anchor is the most label-efficient summary of the past, while the per-period models are individually noisier but carry the drift signal. The delta sequence is
\begin{equation}
d_m = \theta(g_m) - \theta(g_0), \qquad m = 0, \dots, M-1 ,
\end{equation}
and $\theta$ is instantiated per model family so that deltas are commensurable across periods:
\begin{itemize}
\item \emph{Parametric models} (logistic regression, MLP): the flattened weight and bias vector. Per-period models are warm-started from the anchor, so $d_m$ measures displacement from the anchor rather than differences in the optimization path.
\item \emph{Tree ensembles} (XGBoost): the vector of all leaf values. The tree structure is frozen to the anchor's and only leaf values are refreshed per period, so splits index the same leaves throughout and the deltas are aligned by construction.
\item \emph{In-context models} (TabPFN, DR-TabPFN): the \emph{logit field}, the model's class log-probabilities at a fixed set of query points, with $g_m$ conditioned on period $m$'s data as context. This requires no access to parameters, which makes NOMADD applicable to black box models.
\end{itemize}

\subsection{Low-rank temporal factorization and damped forecast}
We stack the deltas into $D \in \mathbb{R}^{M \times P}$, where $P$ is the boundary dimension. Per-period fits on short periods are noisy, and the noise is not concentrated in the directions along which the boundary actually travels. We therefore compress with a truncated singular value decomposition $D \approx U_r \Sigma_r V_r^\top$, which yields $r$ latent temporal trajectories $z_{\cdot k} = (U_r \Sigma_r)_{\cdot k} \in \mathbb{R}^{M}$. Each trajectory is extrapolated $h$ steps beyond the last training period with a damped, regularized linear trend,
\begin{equation}
\hat{z}_{M-1+h,\,k} = \ell_k + \frac{s_k}{1+\lambda}\bigl(\phi + \phi^2 + \dots + \phi^h\bigr),
\end{equation}
where $\ell_k$ and $s_k$ are the level and slope of a least-squares fit to $z_{\cdot k}$, the damping $\phi \in (0,1]$ prevents a short trend from being projected indefinitely, and $\lambda \ge 0$ shrinks the slope. Setting $\phi = 1$ and $\lambda = 0$ recovers ordinary least squares, so the damped forecast strictly generalizes a linear trend. The forecast factors are decoded back through $\Sigma_r V_r^\top$ into a predicted delta $\hat{d}_{M-1+h}$ and applied to the anchor with a shrinkage weight,
\begin{equation}
\hat{\theta}_{M-1+h} = \theta(g_0) + \alpha \, \hat{d}_{M-1+h},
\quad \alpha \in \{0, \tfrac{1}{4}, \tfrac{1}{2}, \tfrac{3}{4}, 1\} .
\end{equation}
Because $\alpha = 0$ deploys the frozen anchor unchanged, declining to extrapolate is always available to the selection procedure and the method can never be forced to move a boundary that history does not justify moving.

\subsection{Configuration selection}
The rank $r \in \{1, 2, \text{full}\}$, damping $\phi \in \{0.4, 0.7, 1.0\}$, regularization $\lambda \in \{0, 1\}$ and shrinkage $\alpha$ are chosen by forward validation \emph{inside the training periods}. For each of the last $V = 3$ training positions $m$, the entire pipeline is rebuilt using only periods $0, \dots, m-1$, including a fresh anchor pooled over those periods only, and the resulting forecast is scored against the labels of period $m$. Those are training labels, so no future information enters at any stage. The configuration with the best mean validation score over the held-out positions is deployed. When fewer than two validation positions exist, $\alpha$ is capped at $0.25$, a fixed low-power rule that limits how far an unvalidated forecast can move the boundary.

\subsection{Model-agnostic election}
\begin{figure}[t]
\centering
\begin{tikzpicture}[
  scale=0.97, transform shape,
  font=\scriptsize,
  box/.style={draw, rounded corners=2pt, align=center, inner sep=3pt, minimum height=18pt},
  arr/.style={-{Stealth[length=5pt]}, thick}]
\node[box, fill=gray!12] (per) {per-period models\\$g_0, g_1, \dots, g_{M-1}$};
\node[box, right=16pt of per, fill=blue!8] (delta) {boundary deltas\\$d_m = \theta(g_m) - \theta(g_0^{\text{pool}})$};
\node[box, right=16pt of delta, fill=blue!8] (svd) {low-rank SVD\\over time (rank $r$)};
\node[box, fill=gray!12, below=16pt of per] (anchor) {anchor $g_0^{\text{pool}}$\\(all periods pooled)};
\node[box, below=16pt of svd, fill=orange!12] (trend) {damped ridge trend\\per latent factor ($\phi, \lambda$)};
\node[box, below=16pt of delta, fill=orange!12] (apply) {anchor $+\ \alpha\,\hat{d}_{\text{future}}$\\$\alpha$ validated, $0$ allowed};
\node[box, below=16pt of apply, fill=green!10] (out) {future-period prediction};
\draw[arr] (per) -- (delta);
\draw[arr] (anchor) -- (delta);
\draw[arr] (delta) -- (svd);
\draw[arr] (svd) -- (trend);
\draw[arr] (trend) -- (apply);
\draw[arr] (anchor.east) -- (apply.west);
\draw[arr] (apply) -- (out);
\end{tikzpicture}
\caption{NOMADD's parameter evolution component. Boundary movement across labeled training periods is measured against a pooled anchor, compressed to a few latent temporal factors, extrapolated with a damped trend, and applied with a validated shrinkage $\alpha$. All hyperparameters are selected by forward validation inside the training periods, and no future labels are used.}
\label{fig:schematic}
\end{figure}
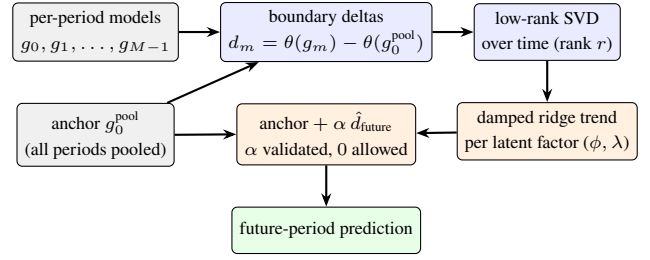

Because the same validation protocol runs unchanged on every base, it also selects \emph{between} bases. We score every candidate on the same held-out training positions and deploy whichever validates best, with the frozen base always among the candidates. A practitioner constrained to a particular model by inference budget or memory can therefore apply the procedure in place, although one can let it make that choice. Section~\ref{sec:headtohead} reports the elected configuration alongside the individual bases. Figure~\ref{fig:schematic} summarizes the parameter evolution procedure.

\section{Experimental Setup}

\noindent
\textbf{Benchmark.} We evaluate on all 18 datasets of the DR-TabPFN benchmark \citep{helli2024drift}, loaded through the authors' own dataset code so that domains, features, and subsampling are identical to their published setup. Fifteen are real datasets and three are synthetic drift streams (Hyperplane, RandomRBF, and rotating 2-Moons) generated by the same families used to pretrain DR-TabPFN. We keep the two groups separate in every aggregate we report, because a method's behavior on a competitor's pretraining distribution is not evidence about its behavior in deployment.

\noindent
\textbf{Protocol.} We use the benchmark's own evaluation protocol rather than a split of our own choosing. For each dataset it selects up to five splits at distinct domain boundaries; for each split the model trains on all preceding domains and is tested on all remaining out-of-distribution domains. Across the 18 datasets this yields 78 valid splits. Scores are pooled one-vs-rest macro ROC-AUC over the out-of-distribution domains, the benchmark's primary metric, computed uniformly for every method from stored predicted probabilities, making our numbers directly comparable to their published table. We report the mean and standard deviation over a dataset's splits, and count a win or loss only outside a $\pm 0.005$ band. For reproducibility: the internal forward validation scores binary datasets by ROC-AUC but multiclass datasets by accuracy, so on multiclass datasets the selection criterion differs from the reporting criterion. All reported numbers are OVR-macro ROC-AUC.

\noindent
\textbf{Methods.} \emph{TabPFN} is the base checkpoint and \emph{DR-TabPFN} is the drift-pretrained checkpoint, the published state of the art. Both run at the paper-default ensemble size $N = 32$ and are called identically, so the checkpoint is the only difference between them. \emph{Ours} is the extrapolation of Section~\ref{sec:method} riding five bases: logistic regression, XGBoost, a small MLP, TabPFN, and DR-TabPFN itself. Every base additionally reports a \emph{frozen} control, which is that same base with $\alpha = 0$ and nothing else changed. Comparing a base against its own frozen control is what isolates the extrapolation from base-model quality, and it is the comparison we lead with.

Two frozen references appear in the tables and are not interchangeable. The \emph{TabPFN} and \emph{DR-TabPFN} baseline columns are pooled fits that receive the true domain index as an input feature, as the benchmark intends. The \emph{frozen} control for our foundation model instantiations is the pooled-context anchor of Section~\ref{sec:anchor} with no domain index, since the extrapolation supplies the temporal information instead. Each comparison uses the reference that makes it a controlled one.

We do not compare against neural temporal domain generalization methods \citep{nasery2021training,bai2023temporal,cai2024continuous,tea2025}: they do not apply to tree ensembles or tabular foundation models, which is the range our claim is about.

\noindent
\textbf{Hardware.} All runs use a single NVIDIA H200 NVL under Linux, with Python 3.12, PyTorch 2.10 and XGBoost 3.2. Random seeds are fixed for NumPy, PyTorch, the trees and the foundation model ensembles. 

\section{Results}

\subsection{What the NOMADD contributes}
\label{sec:contribution}

\begin{figure*}[t]
\centering
\includegraphics[width=0.6\linewidth]{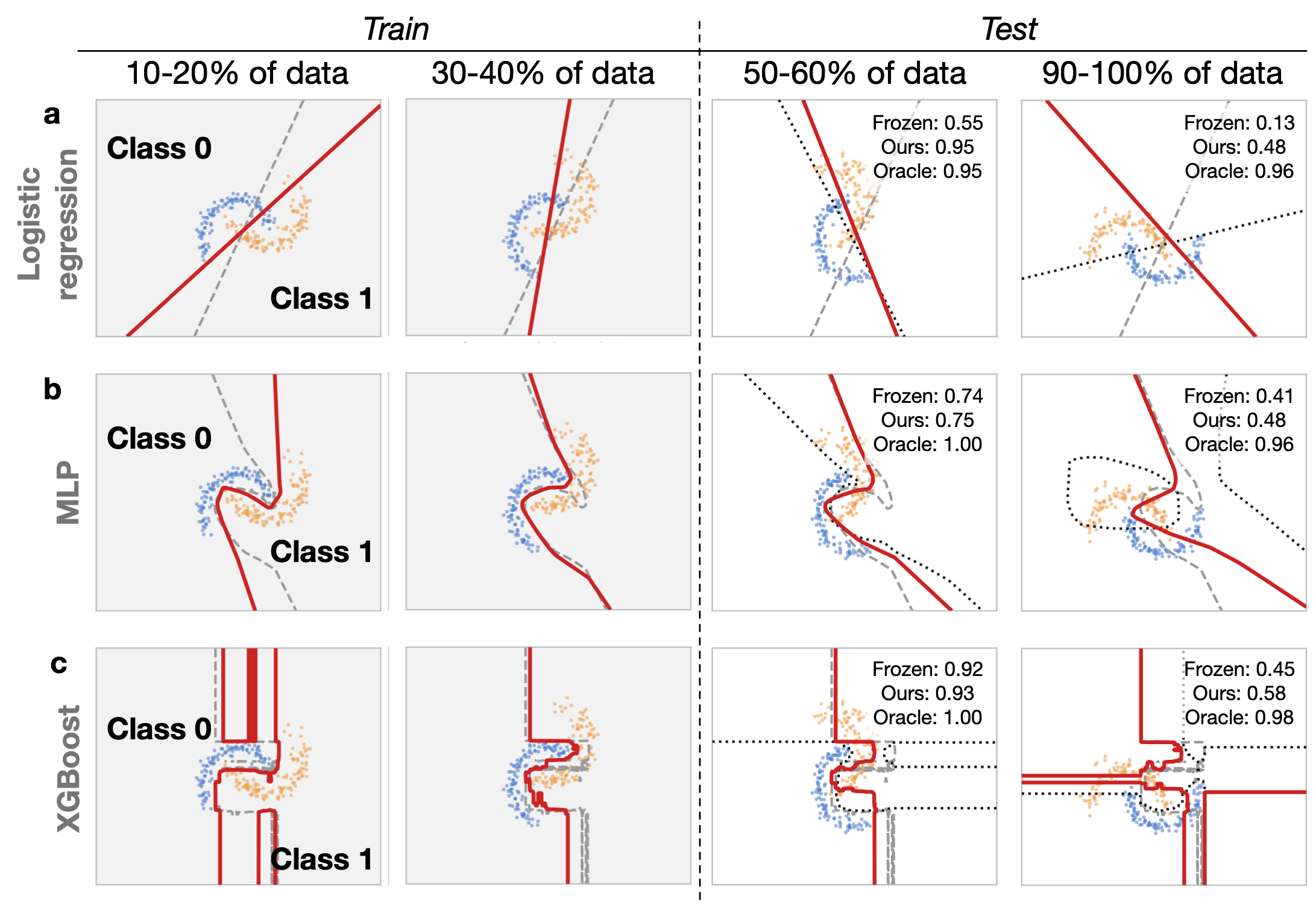}
\caption{Decision-boundary evolution on 2-Moons for three base families across two training and two out-of-distribution test periods. The red line indicates the boundary under NOMADD, gray long-dashed lines indicate the frozen model, and the short-dashed lines indicate the oracle, where a model is trained and tested on that period. We list the model performance (OVR-macro ROC-AUC) on the test periods under each situation as well.
}
\label{fig:boundary_moons}
\end{figure*}

We compare NOMADD-augmented models against the frozen model in Table~\ref{tab:frozen_vs_extended}. Figure~\ref{fig:boundary_moons}, meanwhile, shows how NOMADD performs in practice with an example datasets, 2-Moons. Adding NOMADD gives a positive mean gain on every base family: logistic $+0.030$, XGBoost $+0.028$, MLP $+0.015$, TabPFN $+0.005$, and even DR-TabPFN $+0.007$. The clearest result is for XGBoost where 13 datasets improve, 3 do not significantly change (based on Wilcoxon signed rank test, p-value$>0.05$), and 2 are harmed by NOMADD. 
The two foundation model gains are close to zero, consistent with the redundancy argument above: the value of the method is on lighter models, where NOMADD supplies information the model does not already have. The per-dataset gains are visible in the supplementary material, which plots the change from adding NOMADD for every dataset and every base family. The largest individual improvements fall on the synthetic streams, whose boundaries rotate or translate on a fixed schedule, which is exactly the regime the method assumes. Real datasets drift less regularly and the per-dataset gains there are correspondingly smaller. 

\begin{table*}[t]
\centering
\footnotesize
\setlength{\tabcolsep}{3pt}
%
\begin{tabular}{l|cc|cc|cc|cc|cc}
\toprule
& \multicolumn{2}{c|}{Logistic} & \multicolumn{2}{c|}{XGBoost} & \multicolumn{2}{c|}{MLP} & \multicolumn{2}{c|}{TabPFN} & \multicolumn{2}{c}{DR-TabPFN} \\
Dataset & frozen & +NOMADD & frozen & +NOMADD & frozen & +NOMADD & frozen & +NOMADD & frozen & +NOMADD \\
\midrule
Electricity & \textbf{0.739} & 0.722 & \textbf{0.728} & 0.724 & 0.714 & \textbf{0.721} & \textbf{0.775} & 0.762 & \textbf{0.778} & 0.771 \\
Chess & \textbf{0.751} & 0.748 & \textbf{0.705} & 0.704 & 0.707 & \textbf{0.723} & \textbf{0.741} & \textbf{0.741} & \textbf{0.750} & 0.749 \\
Occupancy & \textbf{0.998} & \textbf{0.998} & \textbf{0.995} & \textbf{0.995} & \textbf{0.944} & 0.920 & \textbf{0.999} & \textbf{0.999} & \textbf{0.999} & \textbf{0.999} \\
Housing-Ames & \textbf{0.839} & 0.814 & \textbf{0.927} & \textbf{0.927} & \textbf{0.787} & 0.729 & 0.957 & \textbf{0.967} & \textbf{0.955} & 0.953 \\
Urban-Traffic & \textbf{0.669} & 0.641 & 0.730 & \textbf{0.744} & 0.694 & \textbf{0.711} & 0.784 & \textbf{0.798} & \textbf{0.802} & 0.801 \\
Istanbul-Stock-Exchange & \textbf{0.717} & 0.716 & 0.677 & \textbf{0.703} & \textbf{0.650} & 0.607 & \textbf{0.798} & 0.741 & \textbf{0.798} & 0.782 \\
Free-Light-Chain & 0.919 & \textbf{0.922} & 0.881 & \textbf{0.885} & 0.845 & \textbf{0.868} & \textbf{0.919} & 0.918 & \textbf{0.922} & \textbf{0.922} \\
Absenteeism & 0.672 & \textbf{0.692} & 0.712 & \textbf{0.715} & 0.646 & \textbf{0.650} & \textbf{0.749} & 0.745 & \textbf{0.757} & \textbf{0.757} \\
Cleveland-Heart & 0.864 & \textbf{0.868} & 0.838 & \textbf{0.843} & 0.822 & \textbf{0.855} & 0.871 & \textbf{0.872} & \textbf{0.843} & 0.842 \\
Parking-Birmingham & 0.787 & \textbf{0.796} & 0.824 & \textbf{0.835} & \textbf{0.815} & 0.797 & \textbf{0.856} & 0.834 & \textbf{0.861} & 0.838 \\
Airlines & \textbf{0.515} & \textbf{0.515} & 0.556 & \textbf{0.560} & \textbf{0.533} & 0.531 & \textbf{0.555} & 0.528 & \textbf{0.548} & 0.522 \\
Diabetes-Questionaire & \textbf{0.907} & 0.875 & 0.907 & \textbf{0.924} & 0.761 & \textbf{0.821} & \textbf{0.898} & 0.873 & \textbf{0.944} & 0.932 \\
Indian-Liver & \textbf{0.769} & 0.743 & 0.702 & \textbf{0.727} & 0.712 & \textbf{0.720} & \textbf{0.790} & 0.784 & \textbf{0.783} & 0.780 \\
Diabetes-Pima & 0.673 & \textbf{0.703} & 0.674 & \textbf{0.730} & 0.564 & \textbf{0.611} & \textbf{0.722} & 0.678 & 0.745 & \textbf{0.750} \\
Diabetic-Hospitals & 0.518 & \textbf{0.520} & 0.526 & \textbf{0.532} & \textbf{0.531} & 0.528 & 0.524 & \textbf{0.530} & \textbf{0.536} & 0.528 \\
\midrule
Hyperplane$^{*}$ & 0.484 & \textbf{0.760} & 0.492 & \textbf{0.711} & 0.504 & \textbf{0.678} & 0.484 & \textbf{0.701} & 0.488 & \textbf{0.643} \\
RandomRBF$^{*}$ & \textbf{0.564} & 0.529 & \textbf{0.903} & \textbf{0.903} & \textbf{0.828} & 0.804 & \textbf{0.859} & 0.841 & \textbf{0.883} & 0.868 \\
2-Moons$^{*}$ & 0.406 & \textbf{0.776} & 0.695 & \textbf{0.808} & 0.587 & \textbf{0.631} & 0.843 & \textbf{0.908} & 0.850 & \textbf{0.926} \\
\midrule
$\Delta$ ROC-AUC, all 18 & \multicolumn{2}{c|}{$+0.030$} & \multicolumn{2}{c|}{$\mathbf{+0.028}$} & \multicolumn{2}{c|}{$+0.015$} & \multicolumn{2}{c|}{$+0.005$} & \multicolumn{2}{c}{$+0.007$} \\
$\Delta$ ROC-AUC, real 15  & \multicolumn{2}{c|}{$-0.004$} & \multicolumn{2}{c|}{$\mathbf{+0.011}$} & \multicolumn{2}{c|}{$+0.004$} & \multicolumn{2}{c|}{$-0.011$} & \multicolumn{2}{c}{$\mathbf{-0.006}$} \\
\bottomrule
\end{tabular}%
\caption{Per-dataset pooled out-of-distribution OVR-macro ROC-AUC, each base family frozen versus NOMADD. The better of each pair is bold. Rows marked $^{*}$ are the synthetic streams. The last two rows give the mean gain over all 18 datasets and over the 15 real datasets alone; a bold mean gain is significant under a Wilcoxon signed-rank test on the datasets in that row ($p < 0.05$). The gain is positive for every family over all 18 datasets, but that aggregate is carried by the synthetic streams: on real data alone only XGBoost gains significantly ($+0.011$, $p = 0.005$), and the significant DR-TabPFN entry is a small \emph{decrease} ($p = 0.010$).}
\label{tab:frozen_vs_extended}
\end{table*}

\subsection{Comparison with DR-TabPFN}
\label{sec:headtohead}

Supplementary Table~1 places every instantiation next to the base checkpoint and DR-TabPFN on all 18 datasets. Averaged over the benchmark, DR-TabPFN leads at 0.817 against 0.798 for our strongest instantiation, which rides the same checkpoint. Restricted to the 15 real datasets the two are much closer, 0.802 against 0.795, a gap of 0.007 that is not significant ($p = 0.069$ paired $t$-test, $p = 0.091$ Wilcoxon), and the base checkpoint sits between them at 0.800; Supplementary Figure~2 gives the per-dataset scatter. The cost picture cuts the other way here: riding the pretrained checkpoint means inheriting its memory footprint, so it is the lightweight bases that are cheaper than DR-TabPFN (Section~\ref{sec:cost}). 

\subsection{Testing NOMADD's Assumptions}
\label{sec:why}

\begin{figure}[tbh!]
\centering
\includegraphics[width=1\linewidth]{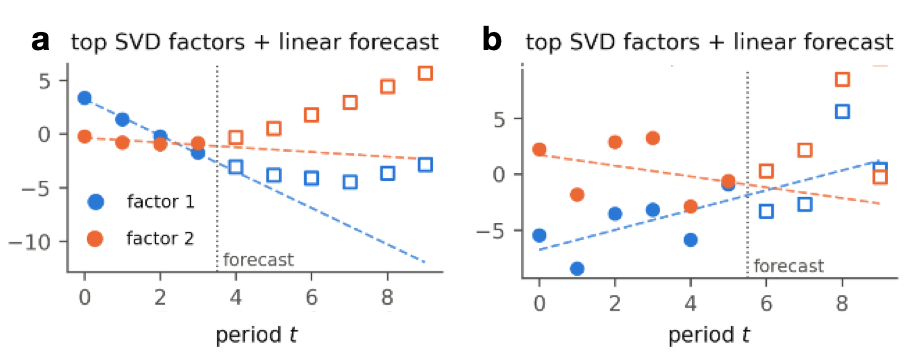}
\caption{Boundary parameter trajectories over time for (a) 2-Moons  and (b) Housing-Ames. The former is exemplar of synthetic data in the DR-TabPFN benchmark, while the latter is exemplar of the real data. The curves are the top two latent factors recovered by the low-rank factorization, with filled circles for training periods, open squares for held-out future periods, and an \emph{undamped} least-squares trend as a dashed line. 
}
\label{fig:params}
\end{figure}

The method rests on two assumptions: that the boundary moves along a small number of directions, and that its motion along those directions is smooth enough to project forward. Figure~\ref{fig:params} shows the top latent factors that the method extrapolates, with an undamped least-squares trend fitted on the training periods and the held-out values marked. Both trajectories are ordered and directional but the plain trend overshoots the held-out values once projected beyond the training range, and it overshoots further on the noisier real dataset. This is the empirical case for the damping $\phi$ and the validated shrinkage $\alpha$: the deployed forecast never applies the raw trend. Supplementary Figure~3 meanwhile shows how the boundary evolves in real data, thus demonstrating the need for models to address concept drift. Finally, Supplementary Figure~4 demonstrates how the synthetic datasets saturate within two factors while real datasets need more, which is why the rank is validated per dataset rather than fixed. 

\subsection{Ablations}
\label{sec:ablation}

Each ablation changes exactly one component and leaves everything else identical, including the validation protocol. Table~\ref{tab:ablation} reports the mean change in pooled ROC-AUC by replacing the deployed choice with the alternative. The ``pulling data back'' component is where we move the test data back to the train data means and covariates if this improves validation performance, therefore only a subset of datasets have this component (and never the foundation models).
This component does not strongly affect model performance, although is mildly helpful for all but the MLP model, so little is lost by the train-only default. Replacing the SVD with the full parameter space (``No SVD'' column) tends to help the model. While unexpected (as choosing SVD or not is determined via validation data in NOMADD) the effect is very small. Regularization of the parameter drift (``No $\alpha$'' column), meanwhile, has a minor impact on model performance, typically improving test performance.

\begin{table}[t]
\centering
\small
\begin{tabular}{lccc}
\toprule
Base & No pull-back$^{\dagger}$ & No SVD & No $\alpha$\\
\midrule
Logistic  & $-0.001$ (11) & $-0.004$ & $-0.005$ \\
XGBoost   & $-0.001$ (10) & $+0.001$ & $-0.002$ \\
MLP       & $+0.019$ (14) & $+0.002$ & $-0.003$ \\
TabPFN    & --            & $+0.010$ & $+0.001$ \\
DR-TabPFN & --            & $+0.000$ & $-0.000$ \\
Elected   & $-0.004$ (17) & $+0.002$ & $+0.005$ \\
\bottomrule
\end{tabular}
\caption{Average increase or decrease in OVR-macro ROC-AUC when NOMADD has each component removed. 
$^{\dagger}$The data pull-back is used only on a subset of datasets, chosen by validation performance, so this column is averaged over those datasets with the count in parentheses; it is never selected on either transformer base, hence the dashes.
}
\label{tab:ablation}
\end{table}

\subsection{Cost: inference time and memory}
\label{sec:cost}

\textbf{Complexity.} NOMADD fits models within each training period. The factorization, trend fit, and decode cost $O(M^2 P + rPM)$ for a boundary of size $P$, and the configuration search reuses the $M$ fitted models rather than refitting them; NOMADD's extrapolation itself takes at most a few milliseconds. Prediction is a single base forward pass, identical in cost to the frozen base, except for the in-context bases, whose logit fields must be re-evaluated for each new batch of test points at a cost of $M$ context passes.

\begin{figure*}[t]
\centering
\includegraphics[width=0.9\linewidth]{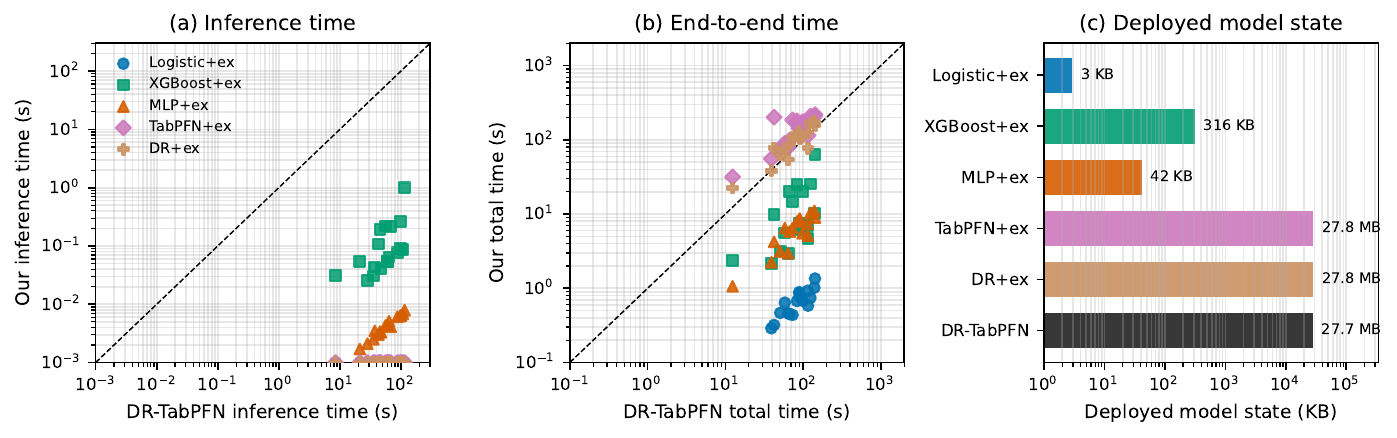}
\caption{Time and memory cost of NOMADD and DR-TabPFN. (a) Prediction time and (b) model training time (or in-context learning time for tabular foundation models) plus NOMADD training, one point per dataset, both on log axes, with the dashed diagonal marking parity. Points below the line represent NOMADD being faster than DR-TabPFN. (c) Mean deployed model size in KB. Values in (a) are floored at the 1\,ms resolution.
}
\label{fig:cost}
\end{figure*}

\textbf{Measured cost.} Table~\ref{tab:cost} and Figure~\ref{fig:cost} report fitting time, prediction time, and peak GPU memory, recorded during the same run that produced every score in this paper on a single NVIDIA H200. Fitting covers all model fits plus the forward-validation search; prediction is measured separately on the out-of-distribution periods. At prediction time the lightweight bases are faster by three to four orders of magnitude: a median of 0.07\,s for XGBoost and 0.004\,s for the MLP against 58.2\,s for DR-TabPFN, with logistic regression below the 1\,ms resolution of our timer. End to end, the logistic regression is a median of $133\times$ faster and the XGBoost and MLP are roughly $11\times$ faster than DR-TabPFN. None of the three allocates any GPU memory, since they run entirely on the CPU, against 135\,MB of resident weights for the foundation models.

There are three caveats to these results. First, DR-TabPFN requires substantial pre-training: millions of synthetic datasets over approximately 1,300 GPU-hours \citep[Appendix A.3]{helli2024drift}. Including that into our figure would make the performance trivially better for lighter models. Next, our tabular foundation model instantiations are \emph{slower} to fit than the model they are compared against (141\,s and 101\,s against 23\,s) because the procedure fits the base once per training period. And their near-zero prediction times in Figure~\ref{fig:cost}(a) are not free: the logit field is evaluated during fitting, so that cost appears in the fitting column. The cost advantage therefore belongs to the lightweight bases, where the accuracy advantage of Section~\ref{sec:contribution} is concentrated as well. This is where NOMADD's model-agnostic property pays: a practitioner constrained to a small inference budget cannot deploy a tabular foundation model at all, so their relevant comparison is an extrapolated XGBoost against a frozen one.

\begin{table}[t]
\centering
\small
\setlength{\tabcolsep}{2.95pt}
\begin{tabular}{@{}lcccc@{}}
\toprule
Method & Fit (s) & Pred.\ (s) & GPU (MB) & Size \\
\midrule
TabPFN & 33.9 & 86.78 & 132 & 27.7 MB \\
DR-TabPFN & 23.0 & 60.14 & 135 & 27.7 MB \\
\midrule
Logistic+NOMADD & \textbf{0.6} & \textbf{0.00} & \textbf{0} & \textbf{3 KB} \\
XGBoost+NOMADD & 12.9 & 0.15 & \textbf{0} & 316 KB \\
MLP+NOMADD & 6.3 & \textbf{0.00} & \textbf{0} & 42 KB \\
TabPFN+NOMADD & 141.1 & 0.00 & 145 & 27.8 MB \\
DR-TabPFN+NOMADD & 100.6 & 0.00 & 151 & 27.8 MB \\
\bottomrule
\end{tabular}
\caption{Mean cost per dataset over the 18 benchmark datasets, measured on a single NVIDIA H200. Fitting includes all per-period fits and the forward-validation configuration search. \emph{Size} is the deployed artifact that must be shipped and held to predict: weights for the logistic and neural bases, leaf values for the tree, and the pretrained checkpoint plus logit field for the in-context bases. The three lightweight bases run entirely on the CPU and allocate no GPU memory. Prediction time for the in-context instantiations is near zero because their logit fields are evaluated during fitting, not because prediction is free.}
\label{tab:cost}
\end{table}

\section{Discussion}

The mean and covariance of streaming data, as well as the parameters of a model trained on those data, evolve in a predictable fashion. We use this observation to reduce data and concept drift with NOMADD: simple regressions allow a model to better predict the future data labels than the frozen model or data alone. The accuracy depends on how regularly the boundary moves: the largest individual gains fall on data streams whose boundary rotates or translates on a schedule. We also notice that, while DR-TabPFN narrowly beats NOMADD performance, DR-TabPFN requires pre-training on millions of synthetic datasets over approximately 1,300 GPU-hours \citep[Appendix A.3]{helli2024drift}. In addition, DR-TabPFN's GPU memory and inference time is far larger than NOMADD. DR-TabPFN's pre-training also needs to be conducted for each new iteration of TabPFN, and only generalizes to tabular foundation models built on structural causal models. Our method, meanwhile, is applicable to a far wider set of models, and its lightweight instantiations predict orders of magnitude faster with minimal memory burdens over the base model (Table~\ref{tab:cost}). We note that in Fig.~\ref{fig:cost}c, the memory is very close to the memory requirements of the underlying model; thus, the algorithm is a lightweight addition.

\subsection{Limitations}
A few aspects of NOMADD limit its performance: it does not alter the parameters of a tabular foundation model, it assumes linear trends in parameter and data space, features are assumed to be sufficient, and we cannot anticipate abrupt data changes. The reason we do not modify foundation models is because they rely on ICL, rather than modifying parameters themselves. Future work could however evaluate how neurons get activated via ICL over time, and instead boost or dampen particular neurons, rather than change the log-odds predictions like the current version of NOMADD. For trainable models, like XGBoost or neural networks, our estimate of boundary movement is only as good as the per-period fits it is built from, and several benchmark datasets have periods of a few dozen rows, which puts a noise floor under everything downstream. The same scarcity limits the model-agnostic election, which must choose among candidates on those same small periods and is the reason the elected column does not reach the best individual base. In addition, the boundary movement is assumed to be linear, but non-linear changes are possible, as shown in Fig.~\ref{fig:params}. We also assume the feature set is fixed and sufficient. In practice, new features may need to be engineered that better capture rapid changes in data. For example, a fraud model may be better at capturing data drift if it includes features that account for seasonality such as changes in purchase behavior compared to a week ago, a month ago, or a year ago. Finally, a fundamental limitation of NOMADD and many other methods that extrapolate trends is that abrupt changes cannot be anticipated by any method that extrapolates a trend.

\subsection{Future work}

To improve model predictions, our simplified linear trends in data or parameter space could be replaced by a curvilinear one or a lightweight learned predictor for each latent factor. In addition, the truncated SVD could be replaced by a richer latent representation such as an autoencoder. We can also extend this work to unstructured modalities such as text, images, or graphs. Memes and language evolve as new topics, facts, and sentiment change (for example, ``algospeak'' in social media \cite{steen2023you}, or meme images that build on each other \cite{seiffert2018memes}). By anticipating this evolution in latent space, we could improve the accuracy of models built on these data. Similarly, NOMADD for knowledge graphs could predict both missing edges and which edges should be added or pruned due to the natural evolution of knowledge, for example that a person's age changes or that a term-limited office holder is no longer in office. This could update a retrieval database automatically without changing the underlying model, improving downstream outputs without manual curation.

\section{Conclusion}

We present a post-hoc method that turns any sequence of period-wise model fits into a forecast of the future decision boundary, with every hyperparameter chosen by forward validation inside the training periods and no future labels used at any stage. Evaluated on the state of the art's own benchmark, protocol, and primary metric, it improves every base family over its own frozen version, most reliably on the lightweight bases, where the gain on a gradient-boosted tree is significant on both paired tests. Applied to a foundation model, it is statistically indistinguishable from the pretrained state of the art on the real datasets, without any pretraining of its own, and with far less GPU memory and inference time. The method's practical value is especially concentrated where drift handling is otherwise unavailable: on ordinary models that are already deployed, and models that cannot be retrained on demand.

\newpage
\bibliography{refs}

\end{document}


\maketitle


This supplement collects the per-dataset results behind the aggregates reported in the main paper. Section~\ref{si:results} gives the full score table and the paired significance tests; Section~\ref{si:gains} breaks the gain from extrapolation down by dataset and by base family; Section~\ref{si:geometry} shows the boundary geometry that the method relies on. All numbers come from the same run, protocol, and metric as the main paper: the Drift-Resilient TabPFN benchmark's own five-split partition, scored as pooled out-of-distribution one-vs-rest macro ROC-AUC.

\section{Full per-dataset results}
\label{si:results}

Table~\ref{tab:main} reports every instantiation next to the two baselines on all 18 datasets, with the real and synthetic groups separated and the two group means given at the foot. Table~\ref{tab:statistical_comparison} gives the paired $t$-test and Wilcoxon signed-rank comparisons of each instantiation against both baselines, which are the tests behind the significance claims made in the main paper.

\begin{table*}[tbh!]
\centering
\resizebox{\textwidth}{!}{%
\begin{tabular}{l|cc|ccccc|c}
\toprule
Dataset & TabPFN & DR-TabPFN & Logistic+NOMADD & XGBoost+NOMADD & MLP+NOMADD & TabPFN+NOMADD & DR-TabPFN+NOMADD & Elected \\
\midrule
Electricity & \textbf{0.772$\pm$0.020} & 0.771$\pm$0.030 & 0.722$\pm$0.037 & 0.724$\pm$0.048 & 0.721$\pm$0.064 & 0.762$\pm$0.021 & 0.771$\pm$0.037 & 0.767 \\
Chess & 0.740$\pm$0.025 & \textbf{0.752$\pm$0.011} & 0.748$\pm$0.037 & 0.704$\pm$0.026 & 0.723$\pm$0.022 & 0.741$\pm$0.023 & 0.749$\pm$0.019 & 0.748 \\
Occupancy & \textbf{0.999$\pm$0.001} & \textbf{0.999$\pm$0.001} & 0.998$\pm$0.002 & 0.995$\pm$0.004 & 0.920$\pm$0.099 & \textbf{0.999$\pm$0.001} & \textbf{0.999$\pm$0.001} & 0.998 \\
Housing-Ames & 0.955$\pm$0.023 & 0.953$\pm$0.020 & 0.814$\pm$0.081 & 0.927$\pm$0.017 & 0.729$\pm$0.104 & \textbf{0.967$\pm$0.015} & 0.953$\pm$0.025 & 0.853 \\
Urban-Traffic & \textbf{0.862$\pm$0.137} & 0.840$\pm$0.175 & 0.641$\pm$0.100 & 0.744$\pm$0.167 & 0.711$\pm$0.127 & 0.798$\pm$0.178 & 0.801$\pm$0.180 & 0.799 \\
Istanbul-Stock-Exchange & \textbf{0.795$\pm$0.023} & 0.791$\pm$0.023 & 0.716$\pm$0.037 & 0.703$\pm$0.056 & 0.607$\pm$0.057 & 0.741$\pm$0.112 & 0.782$\pm$0.041 & 0.766 \\
Free-Light-Chain & \textbf{0.923$\pm$0.002} & 0.919$\pm$0.004 & 0.922$\pm$0.001 & 0.885$\pm$0.009 & 0.868$\pm$0.033 & 0.918$\pm$0.010 & 0.922$\pm$0.005 & 0.922 \\
Absenteeism & 0.748$\pm$0.010 & 0.748$\pm$0.019 & 0.692$\pm$0.013 & 0.715$\pm$0.014 & 0.650$\pm$0.023 & 0.745$\pm$0.011 & \textbf{0.757$\pm$0.019} & \textbf{0.757} \\
Cleveland-Heart & 0.865$\pm$0.005 & 0.841$\pm$0.004 & 0.868$\pm$0.002 & 0.843$\pm$0.012 & 0.855$\pm$0.015 & \textbf{0.872$\pm$0.006} & 0.842$\pm$0.005 & 0.870 \\
Parking-Birmingham & 0.846$\pm$0.049 & \textbf{0.870$\pm$0.036} & 0.796$\pm$0.036 & 0.835$\pm$0.041 & 0.797$\pm$0.033 & 0.834$\pm$0.061 & 0.838$\pm$0.036 & 0.830 \\
Airlines & 0.552$\pm$0.051 & 0.535$\pm$0.034 & 0.515$\pm$0.012 & \textbf{0.560$\pm$0.039} & 0.531$\pm$0.006 & 0.528$\pm$0.015 & 0.522$\pm$0.007 & 0.526 \\
Diabetes-Questionaire & 0.907$\pm$0.010 & \textbf{0.953$\pm$0.014} & 0.875$\pm$0.018 & 0.924$\pm$0.020 & 0.821$\pm$0.060 & 0.873$\pm$0.026 & 0.932$\pm$0.007 & 0.878 \\
Indian-Liver & 0.782$\pm$0.036 & \textbf{0.784$\pm$0.022} & 0.743$\pm$0.064 & 0.727$\pm$0.033 & 0.720$\pm$0.067 & \textbf{0.784$\pm$0.040} & 0.780$\pm$0.042 & 0.763 \\
Diabetes-Pima & 0.724$\pm$0.013 & 0.744$\pm$0.011 & 0.703$\pm$0.015 & 0.730$\pm$0.002 & 0.611$\pm$0.024 & 0.678$\pm$0.071 & \textbf{0.750$\pm$0.010} & 0.693 \\
Diabetic-Hospitals & 0.526$\pm$0.019 & 0.531$\pm$0.022 & 0.520$\pm$0.015 & \textbf{0.532$\pm$0.015} & 0.528$\pm$0.015 & 0.530$\pm$0.016 & 0.528$\pm$0.024 & 0.526 \\
\midrule
Hyperplane$^{*}$ & 0.729$\pm$0.145 & \textbf{0.801$\pm$0.127} & 0.760$\pm$0.124 & 0.711$\pm$0.103 & 0.678$\pm$0.088 & 0.701$\pm$0.092 & 0.643$\pm$0.123 & 0.760 \\
RandomRBF$^{*}$ & 0.754$\pm$0.061 & 0.901$\pm$0.018 & 0.529$\pm$0.013 & \textbf{0.903$\pm$0.038} & 0.804$\pm$0.034 & 0.841$\pm$0.029 & 0.868$\pm$0.032 & 0.858 \\
2-Moons$^{*}$ & 0.840$\pm$0.124 & \textbf{0.969$\pm$0.031} & 0.776$\pm$0.153 & 0.808$\pm$0.097 & 0.631$\pm$0.097 & 0.908$\pm$0.044 & 0.926$\pm$0.039 & 0.906 \\
\midrule
Mean (real 15) & 0.800 & 0.802 & \textbf{0.752} & \textbf{0.770} & \textbf{0.719} & \textbf{0.785} & 0.795 & \textbf{0.780} \\
Mean (all 18) & 0.795 & 0.817 & \textbf{0.741} & \textbf{0.776} & \textbf{0.717} & \textbf{0.790} & \textbf{0.798} & \textbf{0.790} \\
\bottomrule
\end{tabular}%
}
\caption{Pooled out-of-distribution OVR-macro ROC-AUC on the Drift-Resilient TabPFN benchmark under its own five-split protocol, mean and standard deviation over each dataset's splits. In the per-dataset rows, bold marks the best score in that row across all eight methods, with ties at the reported precision bolded together; baselines and our instantiations are treated alike. Rows marked $^{*}$ are synthetic streams generated by the families used to pretrain Drift-Resilient TabPFN and are excluded from the real-15 mean. The Elected column is the model-agnostic selection of Section 3.5. In the two mean rows bold carries a different meaning: there it marks a method whose difference from Drift-Resilient TabPFN is statistically significant under a Wilcoxon signed-rank test over the datasets in that row ($p < 0.05$), with Drift-Resilient TabPFN itself the reference and therefore unmarked. On the 15 real datasets the two unmarked entries, the base checkpoint ($p = 0.807$) and our strongest instantiation ($p = 0.091$), are the methods that cannot be distinguished from it; over all 18 only the base checkpoint remains indistinguishable ($p = 0.179$), because the three synthetic streams are drawn from Drift-Resilient TabPFN's own pretraining families.}
\label{tab:main}
\end{table*}

\begin{table*}[tbh!]
\centering
\footnotesize
\setlength{\tabcolsep}{3pt}
\begin{tabular}{llccc}
\toprule
Our method & Baseline & Mean diff & $t$-test $p$ & Wilcoxon $p$ \\
\midrule
Logistic+NOMADD  & TabPFN    & $-0.055$ & $0.005$ & $0.001$ \\
Logistic+NOMADD  & DR-TabPFN & $-0.076$ & $0.004$ & $0.000$ \\
XGBoost+NOMADD   & TabPFN    & $-0.019$ & $0.147$ & $0.021$ \\
XGBoost+NOMADD   & DR-TabPFN & $-0.041$ & $0.001$ & $0.001$ \\
MLP+NOMADD       & TabPFN    & $-0.079$ & $0.000$ & $0.000$ \\
MLP+NOMADD       & DR-TabPFN & $-0.100$ & $0.000$ & $0.000$ \\
TabPFN+NOMADD    & TabPFN    & $-0.005$ & $0.543$ & $0.325$ \\ 
TabPFN+NOMADD    & DR-TabPFN & $-0.027$ & $0.006$ & $0.003$ \\
DR-TabPFN+NOMADD & TabPFN    & $+0.002$ & $0.831$ & $0.899$ \\
DR-TabPFN+NOMADD & DR-TabPFN & $-0.019$ & $0.051$ & $0.012$ \\
Elected       & TabPFN    & $-0.006$ & $0.606$ & $0.417$ \\
Elected       & DR-TabPFN & $-0.027$ & $0.003$ & $0.002$ \\
\bottomrule
\end{tabular}
\caption{Paired comparisons across the 18 datasets against the two baselines. Extrapolation on the drift-pretrained base ties the base checkpoint (difference $+0.002$, $p = 0.83$) and is within 0.019 of Drift-Resilient TabPFN, which is not significant under the $t$-test at the 0.05 level. On the real-15 subset that gap narrows to 0.007 ($p = 0.069$).}
\label{tab:statistical_comparison}
\end{table*}

\section{Per-dataset gains from extrapolation}
\label{si:gains}

\begin{figure*}[t]
\centering
\includegraphics[width=0.95\textwidth]{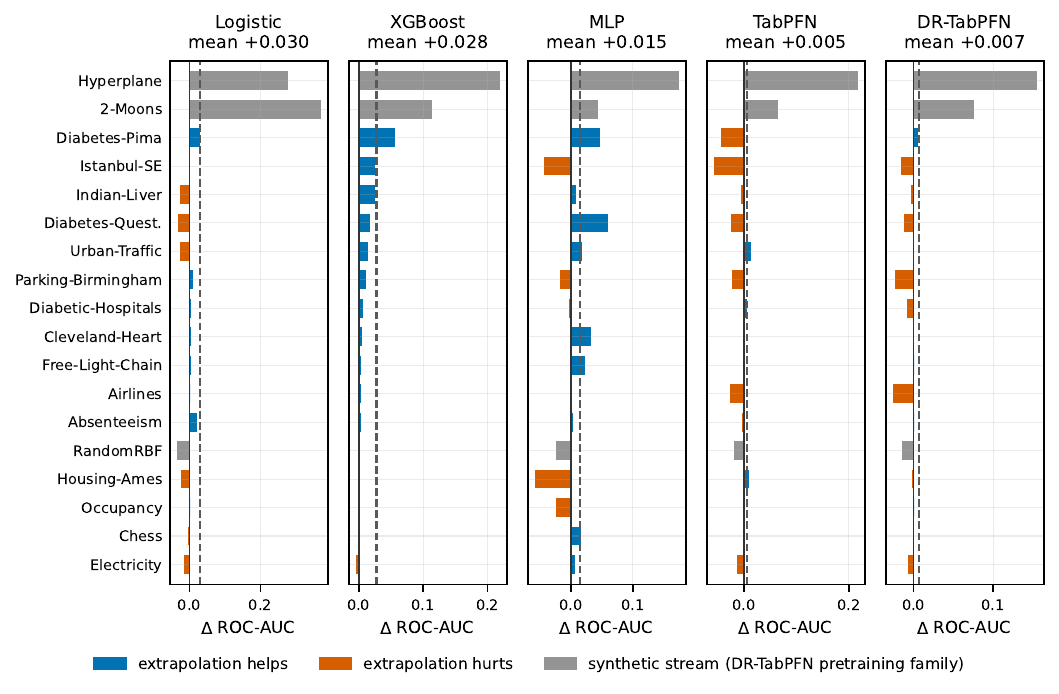}
\caption{Per-dataset change in pooled OVR-macro ROC-AUC from adding extrapolation to each base, against that base left frozen. Datasets are ordered by the XGBoost gain, and the dashed line is each family's mean, which is positive for all five. Grey bars are the three synthetic streams drawn from Drift-Resilient TabPFN's pretraining generators, where the boundary drifts on a fixed schedule and the largest individual gains occur.}
\label{fig:improvement}
\end{figure*}

\begin{figure*}[t]
\centering
\includegraphics[width=0.92\textwidth]{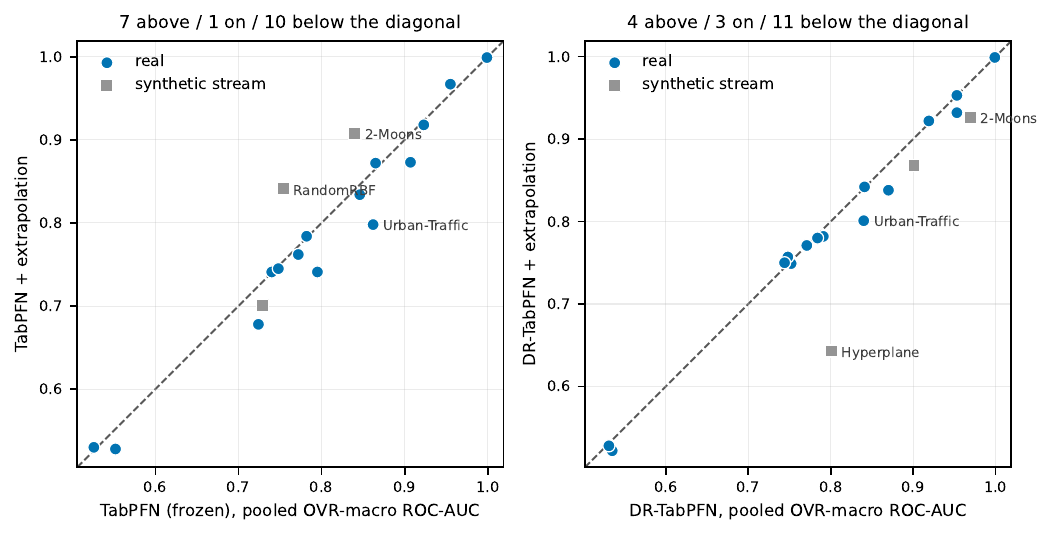}
\caption{Each transformer base with extrapolation against the corresponding frozen control, one point per dataset, pooled OVR-macro ROC-AUC. Points above the dashed diagonal favor extrapolation. The three largest movers in either direction are labeled. 
}
\label{fig:scatter}
\end{figure*}

Figure~\ref{fig:improvement} is the per-dataset breakdown referred to in the main paper: for each base family it plots the change from adding extrapolation, dataset by dataset, against that base left frozen. Figure~\ref{fig:scatter} shows the same comparison for the two transformer bases as a scatter against the frozen control, which makes the split between the synthetic streams and the real datasets easier to read.

\section{Boundary geometry}
\label{si:geometry}

\begin{figure*}[tbh!]
\centering
\includegraphics[width=0.95\linewidth]{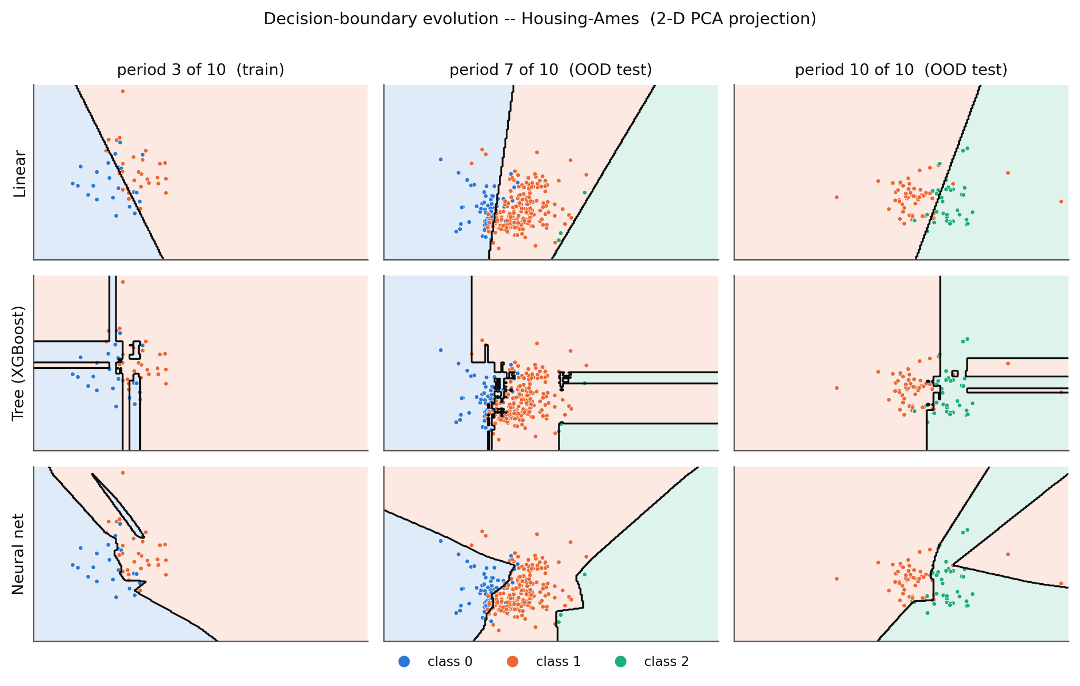}
\caption{Decision-boundary evolution on Housing-Ames, a real three-class dataset, shown in a two-dimensional PCA projection of the feature space. Each panel fits the ``oracle'' model on that period alone in order to display the true boundary at that time. This does not show NOMADD's results; instead it shows how data drift occurs in real data.
}
\label{fig:boundary_housing}
\end{figure*}

\begin{figure}[tbh!]
\centering
\includegraphics[width=0.95\linewidth]{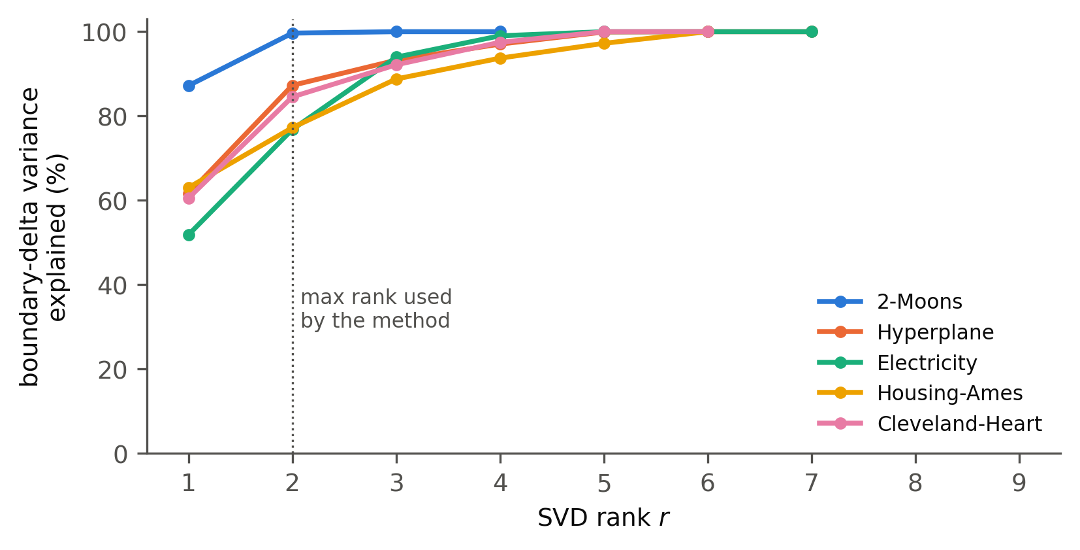}
\caption{Share of boundary movement captured against the rank retained by the factorization. The synthetic streams saturate within two factors; real datasets require more, so the rank is validated per dataset rather than fixed.}
\label{fig:svd}
\end{figure}

The method assumes the boundary moves along a small number of directions and moves smoothly enough to project forward. Figure~\ref{fig:boundary_housing} is the real-data counterpart to the 2-Moons panel in the main paper, showing how the boundary moves on Housing-Ames. Figure~\ref{fig:svd} quantifies the low-rank assumption directly: the synthetic streams saturate within two factors while real datasets need more, which is why the rank is validated per dataset rather than fixed.

